\documentclass{ifacconf}

\usepackage{graphicx}      
\usepackage{natbib}        
\usepackage{amsmath} 
\usepackage{amssymb}  
\usepackage{subfigure}
\usepackage{multirow}
\usepackage{color}
\usepackage{cases}
\newcommand{\argmax}{\mathop{\rm arg~max}\limits}

\begin{document}
\begin{frontmatter}

\title{Learning Locally, Communicating Globally: Reinforcement Learning of Multi-robot Task Allocation for Cooperative Transport}


\author[First]{Kazuki Shibata} 
\author[Second]{Tomohiko Jimbo}
\author[Second]{Tadashi Odashima}
\author[Second]{Keisuke Takeshita} 
\author[Third]{Takamitsu Matsubara}

\address[First]{Applied Mathematics Research-Domain, Toyota Central R$\&$D Labs., Inc., 41-1, Yokomichi, Nagakute, Aichi 480-1192, Japan \\(e-mail: kshibata@mosk.tytlabs.co.jp).}
\address[Second]{R-Frontier Division, Frontier Research Center, Toyota Motor Corporation, 1, Toyota-cho, Toyota, Aichi 471-8571, Japan}
\address[Third]{Division of Information Science, Graduate School of Science and Technology, Nara Institute of Science and Technology, Nara 630-0192, Japan}

\thanks{“This work has been submitted to IFAC for possible publication”.}%

\begin{abstract}                
We consider task allocation for multi-object transport using a multi-robot system, in which each robot selects one object among multiple objects with different and unknown weights. The existing centralized methods assume the number of robots and tasks to be fixed, which is inapplicable to scenarios that differ from the learning environment. Meanwhile, the existing distributed methods limit the minimum number of robots and tasks to a constant value, making them applicable to various numbers of robots and tasks. However, they cannot transport an object whose weight exceeds the load capacity of robots observing the object. To make it applicable to various numbers of robots and objects with different and unknown weights, we propose a framework using multi-agent reinforcement learning for task allocation. First, we introduce a structured policy model consisting of 1) predesigned dynamic task priorities with global communication and 2) a neural network-based distributed policy model that determines the timing for coordination. The distributed policy builds consensus on the high-priority object under local observations and selects cooperative or independent actions. Then, the policy is optimized by multi-agent reinforcement learning through trial and error. This structured policy of local learning and global communication makes our framework applicable to various numbers of robots and objects with different and unknown weights, as demonstrated by numerical simulations.
\end{abstract}

\vspace{-3mm}


\end{frontmatter}

\section{Introduction}
\vspace{-2mm}
In recent years, multi-robot transport has attracted attention in robotics for various applications such as delivery services, factory logistics, and search and rescue. To transport multiple objects over large areas, a team of robots can outperform a single robot in terms of load capacity, time efficiency, and robustness to individual robot failures. Unlike single-robot transport, multi-robot transport involves task allocation and cooperative manipulation. Each robot should select an object to transport multiple objects efficiently. Moreover, force control is required when various robots cooperate to transport a common object to its desired position (\cite{Culbertson2018}).

We consider task allocation for multi-object transport using a multi-robot system. In this study, a task corresponds to an object. The existing studies on multi-robot task allocation have adopted deterministic optimization methods (\cite{Liu2011, Sabattini2017}) or auction methods (\cite{Braquet2021}) under the assumption that the number of robots to execute each task is available. However, these assumptions are not always realistic.
For instance, by using a camera, it may be possible to obtain information on the shape of an object; however, it is challenging to obtain the number of robots required to transport it. In this case, the assumption does not hold.

We explore multi-agent reinforcement learning (MARL) for multi-object transport using a multi-robot system. Each robot selects one object among multiple objects with different and unknown weights. The objective is to transport all the objects to the desired positions as quickly as possible. The existing centralized methods assume the number of robots and tasks to be fixed (\cite{Qie2019, Niwa2022}), which is inapplicable to the scenarios in which the number of robots and tasks differs from the learning environment. Meanwhile, the existing distributed methods limit the minimum number of robots and tasks to a constant value, making them applicable to various numbers of robots and tasks (\cite{Hsu2021}). However, they cannot transport an object whose weight exceeds the load capacity of robots observing the object.

\begin{figure}[!tp]
\centering
\subfigure[Cooperative action]{
\includegraphics[width=3.2cm, height=3.5cm]{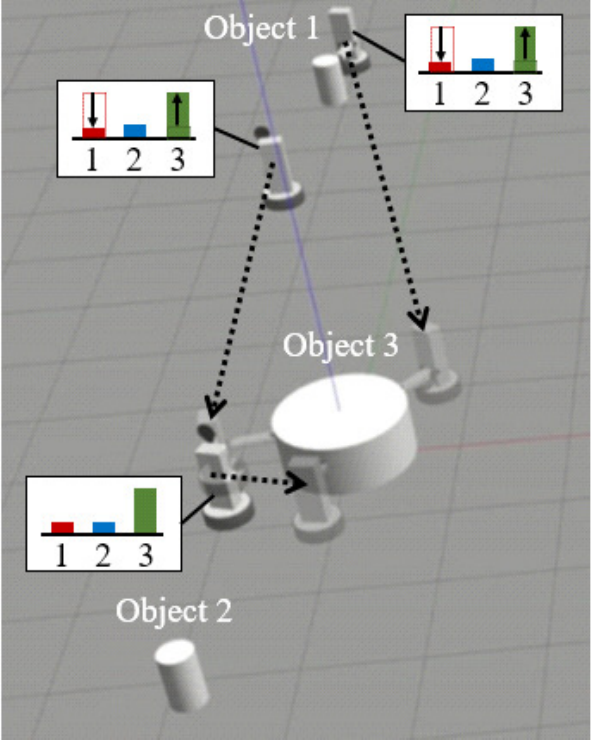}
\label{coop}}
\hspace{2mm}
\subfigure[Independent action]{
\includegraphics[width=3.2cm, height=3.5cm]{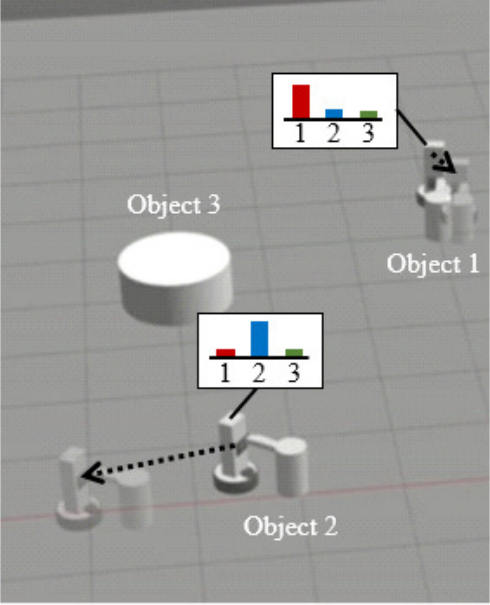}
\label{indi}}
\caption{Multi-object transport using a multi-robot system. (a) Robots perform cooperative actions by building a consensus on the high-priority object when they cannot move the object. (b) Robots perform independent actions when they can move the selected objects.}
\label{fig1}
\end{figure}

To utilize the advantages of the centralized and distributed methods, we propose a framework using the MARL for task allocation. The proposed framework first uses a structured policy model consisting of 1) predesigned dynamic task priorities with global communication and 2) a neural-network-based distributed policy model that determines the timing for coordination. The distributed policy reaches a consensus regarding high-priority tasks under local observations and selects cooperative or independent actions, as illustrated in Fig. \ref{fig1}. The policy is optimized by the MARL through trial and error. This structured policy of local learning and global communication makes our framework suitable for scenarios where the numbers of robots and objects vary, and the number of robots required to transport an object is unknown. Results from the multi-object transport simulations demonstrate that, compared to other methods, our framework can reduce the transport time while transporting all the objects to the desired positions for various numbers of robots and objects.

The contributions of this study can be summarized as follows:
\begin{itemize}
\item We propose a learning framework using a structured policy model consisting of predesigned dynamic task priorities with global communication and a neural-network-based distributed policy model for multi-robot task allocation.
\item Unlike the deterministic optimization and auction methods, our method does not require the number of robots to execute each task and can be applied to a wide range of task allocation problems.
\item We confirm that our method can maintain the high performance for various numbers of robots and objects with different and unknown weights through multi-object transport simulations.
\end{itemize}

The remainder of this paper is organized as follows. Section 2 presents the related work on multi-robot task allocation. Section 3 describes the allocation problem for multi-object transport using a team of robots. Section 4 details the MARL and the proposed learning framework. Section 5 shows the effectiveness of our framework through multi-robot transport simulations. Finally, section 6 summarizes the study and provides directions for future work.

\section{RELATED WORK}
\subsection{Deterministic Optimization Methods}
\vspace{-1mm}
Deterministic optimization formulates the task allocation problem as an optimization problem aimed at minimizing the total travel distance under constraints for the number of robots required for each task. These approaches have adopted various optimization techniques, such as the Hungarian algorithm (\cite{Liu2011}) and integer linear programming (\cite{Sabattini2017}).
Although these studies can guarantee optimality in terms of the total travel distance, most methods require prior information regarding the number of robots required for each task.

\subsection{Distributed Metaheuristic Methods}
\vspace{-1mm}
Metaheuristic methods are inspired by the division of labor exhibited by social insects. A common approach has adopted threshold models (\cite{Theraulaz1998, Krieger2000}), in which each robot selects a task under local observations using an activation threshold and a stimulus associated with each task. Although these methods can handle varying numbers of robots and tasks, they may allocate unnecessary tasks to robots, thus reducing the time efficiency.

\vspace{-1mm}

\subsection{Auction Methods}
\vspace{-1mm}
Auction algorithms (\cite{Gerkey2004}) are common methods for multi-robot task allocation and have been studied in the centralized and decentralized approaches. The centralized method (\cite{Kwasnica2005}) adopts the auctioneer, which collects the bids from the bidders, and allocates the highest bidder to the task. In contrast, \cite{Choi2009} proposes a decentralized auction-based algorithm without the auctioneer. This method adopts a consensus algorithm to estimate the bids of other robots. Then, the robots allocate the task to the highest bidder using the estimated bids. Therefore, each robot can assign a task even if it can locally communicate with other robots. However, their method focuses on the problem where a single robot can execute each task.

\cite{Braquet2021} addressed the closest problem to our study, where each task requires multiple robots. Their method adopts the consensus algorithm similar to \cite{Choi2009}, which estimates the list of selected tasks, the list of winning bids, and the list of completed allocations. Robots assign a task to the robot with the highest bid among the unassigned robots based on the list of completed allocations. Therefore, their method can be applicable to the problem where each task requires multiple robots. However, their methods require a probability of completing each task, which is difficult to compute for objects with unknown weights.

\vspace{-1mm}

\subsection{MARL Methods}
\vspace{-1.5mm}
Recent studies (\cite{Qie2019, Niwa2022}) have addressed task allocation problems using the MARL. These approaches formulate a task allocation problem using the Markov decision process and learn the optimal policies using a multi-agent deep deterministic policy gradient (MADDPG) (\cite{Lowe2017}). However, these methods adopt centralized training assuming that the number of robots and tasks is constant, failing in scenarios with different numbers of robots and tasks.
To address this problem, \cite{Hsu2021} proposed a distributed policy model, which limits the minimum number of robots and tasks to be constant. The trained policies are applicable to up to 1000 robots and 1000 tasks through multi-target tracking simulations. Although their methods can be applied to various numbers of robots and tasks, they cannot handle a situation where the number of robots required to execute a task exceeds the number of robots observing it.

Although the proposed framework uses distributed policies under local observations, it differs from the method (\cite{Hsu2021}) in that our method employs a structured policy model consisting of predesigned dynamic task priorities with global communication and a neural network-based distributed policy model. Therefore, robots can perform all the tasks efficiently even when the number of robots required to complete each task is different and unknown.

\section{Preliminary}
\subsection{Problem Formulation}
We consider a team of $N$ robots. Each of these robots selects one object simultaneously among the $M$ objects with different and unknown weights.
The position of robot $i\left(i=1,\cdots,N\right)$ is represented by $\textit{\textbf{x}}_i\in \mathbb{R}^2$.
The position, velocity, and desired position of the object $l\left(l=1,\cdots,M\right)$ are represented by $\textit{\textbf{z}}_l\in \mathbb{R}^2$, $\textit{\textbf{v}}_l\in \mathbb{R}^2$ and $\textit{\textbf{z}}_l^{\ast}\in \mathbb{R}^2$, respectively.
Robot $i$ can observe robots $j\in \mathcal{N}_i^{\rm Robot}$ and objects $l\in \mathcal{N}_i^{\rm Load}$, whose positions are $K$ nearest from $\textit{\textbf{x}}_i$.
In this study, we simplify the transport problem such that the robots can move the object if the total load capacity of the robot within a certain distance from the object exceeds the mass of the object.

The objective is to transport all the objects to the desired positions as quickly as possible.

We made the following assumptions:
\begin{itemize}
\item Robots know $M$ and $N$
\item Robots know the current and desired positions of $M$ objects
\item Robots can communicate with other robots if necessary
\end{itemize}

\vspace{-1mm}

\subsection{MARL Settings for Multi-robot Task Allocation}
To address the multi-robot task allocation problem for multi-object transport, we describe the MARL settings using a Markov decision process.

Let us denote the state, action, and observation of robot $i$ ($i=1,\cdots,N$) as $\textit{\textbf{s}}_i$, $\textit{\textbf{a}}_i$, and $\textit{\textbf{o}}_i$, respectively.
Robot $i$ selects action $\textit{\textbf{a}}_i$ under local observation $\textit{\textbf{o}}_i$ including robots $j\in \mathcal{N}_i^{\rm Robot}$ and objects $l\in \mathcal{N}_i^{\rm Load}$.
Action $\textit{\textbf{a}}_i$ includes a variable to compute the priority of objects $l\in \mathcal{N}_i^{\rm Load}$ and variables to determine communicating task priorities with other robots, as described in Section 4.
Robot $i$ updates the task priorities by computing the current actions $\textit{\textbf{a}}_i$, then selects the object with the highest priority among the $M$ objects.
After robot $i$ moves to the selected object for a certain control period, $\textit{\textbf{s}}_i$ transitions to the next state $\textit{\textbf{s}}'_i$.
Simultaneously, robot $i$ receives reward $r_t$ at every step $t$ when moving the object or carrying it to the desired position.
Robot $i$ updates its policy by maximizing the expected reward $\mathbb{E}[R_t]=\mathbb{E}\left[\sum^{T-1}_{k=0}\gamma^{k}r_{t+k}\right]$, where $\gamma \in [0,1]$ is a discount factor and $T$ is the total number of steps per episode.

\vspace{-2mm}

\section{METHOD}
In this section, we introduce the proposed MARL framework that can handle a varying number of robots and objects with different and unknown weights.

\begin{figure}[!tp]
\begin{center}
\includegraphics[width=8cm]{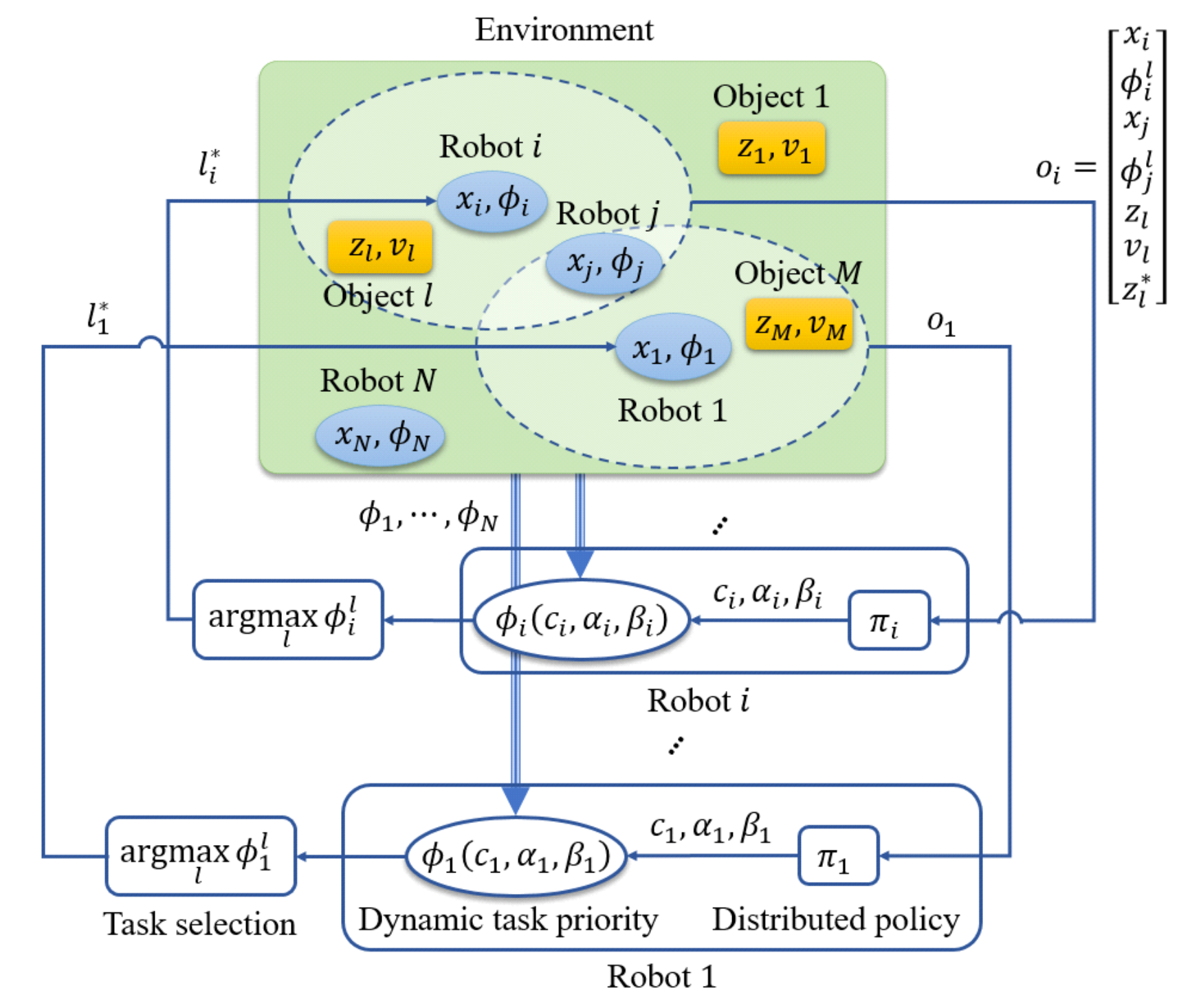}
\vspace{-2mm}
\caption{Overview of learning framework. Robot $i$ updates task priorities of the neighboring objects using $\textit{\textbf{c}}_i$ while building consensus on the high-priority object using $\alpha_i$ and $\beta_i$ according to the distributed policy $\pi_i$ under local observations $\textit{\textbf{o}}_i$. Robot $i$ selects the object $l_i^{\ast}$ which has the highest priority among $M$ objects.}
\label{framework}
\end{center}
\end{figure}

Fig. \ref{framework} shows the overview of the learning framework.
Robot $i$ has task priority $\boldsymbol{\phi}_i:=\left[\phi_i^1,\cdots,\phi_i^M\right]^{\top}\in \mathbb{R}^M$, where $\phi_i^l\in [0,1]$ is the priority of the $l$th object possessed by robot $i$. Robot $i$ updates the priority of the neighboring object $l\in \mathcal{N}_i^{\rm Load}$ under local observation $\textit{\textbf{o}}_i=\left[\textit{\textbf{x}}_i, \phi_i^{l},\textit{\textbf{x}}_{j},\phi_{j}^{l},\textit{\textbf{z}}_{l},\textit{\textbf{v}}_{l},\textit{\textbf{z}}_{l}^{\ast}\right]$ using action $\textit{\textbf{c}}_i=\left[c_i^1,\cdots,c_i^K\right]^{\top}\in \mathbb{R}^K$, where $c_i^l\in[0,1]$ is the reference value of $\phi_i^l$.
Limiting the minimum number of robots and objects to a constant value makes the policy applicable to varying numbers of robots and objects.
However, this policy cannot transport an object whose weight exceeds the load capacity of robots observing the object because it cannot update the priorities of the object $l\notin \mathcal{N}_i^{\rm Load}$.

The proposed framework introduces dynamic task priorities with global communication and a neural network-based distributed policy model.
The distributed policy computes communication inputs $\alpha_i\in [0,1]$ and $\beta_i\in [0,1]$ under local observations, where $\alpha_i$ is the parameter by which the robot $i$ receives task priority from other robots, and $\beta_i$ is the parameter by which the robot $i$ sends $\boldsymbol{\phi}_i$ to other robots.
If robots communicate the task priority with other robots, the dynamic task priority makes the agents establish a consensus on the high-priority object and select cooperative actions. Otherwise, robots select independent actions. Therefore, robots can transport all objects efficiently without knowing the number of robots required to transport objects.
Robot $i$ selects the object $l_i^{\ast}$, which has the highest priority among $M$ objects.
Then, the policy is optimized by MARL through trial and error.

\vspace{-3mm}

\subsection{Dynamic Task Priority with Global Communication}
This subsection introduces the dynamic task priority with global communication to select an object among various candidates.

We design the dynamic task priority such that the robot $i$ can update $\phi_i^l$ ($l\in \mathcal{N}_i^{\rm Load}$) according to its policy while updating $\phi_i^l$ ($l\notin \mathcal{N}_i^{\rm Load}$) using the priorities of the $N$ robots.
In this case, the robots should balance cooperative and independent actions to transport all the objects efficiently.
To this end, we design the dynamic task priority of object $l$ for robot $i$ given by
\begin{eqnarray}
\dot{\phi}_i^l=
\begin{cases}
k_{\phi}(c_i^l-\phi_i^l)+\sigma_i\sum_{j=1}^N d_j k_{\phi}(\phi_j^l-\phi_i^l),\ {\rm if}\ l\in \mathcal{N}_i^{\rm Load}\\
\sigma_i\sum_{j=1}^N d_j k_{\phi}(\phi_j^l-\phi_i^l), {\rm otherwise}
\end{cases}
\label{priority}
\end{eqnarray}
where $k_{\phi}>0$, $d_i$ and $\sigma_i$ are equal to 0 or 1.
We introduced the first-order linear time-delay system to avoid the occurrences of chattering, where the robots travel back and forth between different objects.
$k_{\phi}(c_i^l-\phi_i^l)$ induces an independent action while $k_{\phi}(\phi_j^l-\phi_i^l)$ induces a cooperative action.
If $\sigma_i=1$ and $d_j=1$, $k_{\phi}(\phi_j^l-\phi_i^l)$ makes $\phi_i^l$ asymptotically converge to $\phi_j^l$, establishing consensus on the task priority.
Otherwise, $k_{\phi}(c_i^l-\phi_i^l)$ makes $\phi_i^l$ asymptotically converge to $c_i^l$ according to its own policy.
The distributed policy calculates $\sigma_i$ and $d_i$ to reach a consensus on the high-priority object as well as $\textit{\textbf{c}}_i$ under local observations.

\vspace{-1mm}

\subsection{Distributed Policy Model}
\vspace{-1mm}
We introduce a distributed policy model under local observations $\textit{\textbf{o}}_i$ given by
\begin{eqnarray}
\textit{\textbf{a}}_i=\left[\textit{\textbf{c}}_i^{\top},\alpha_i,\beta_i\right]^{\top}=\pi_i(\textit{\textbf{o}}_i),
\label{policy}
\end{eqnarray}
where $\pi_i$ is computed by a deep neural network.
Agent $i$ determines the reference values of $\phi_i^l$ using $c_i^l$ for $K$ local objects while maintaining the priority of the $M$ objects.

\begin{figure}[!tp]
\vspace{-2mm}
\begin{center}
\includegraphics[width=7cm]{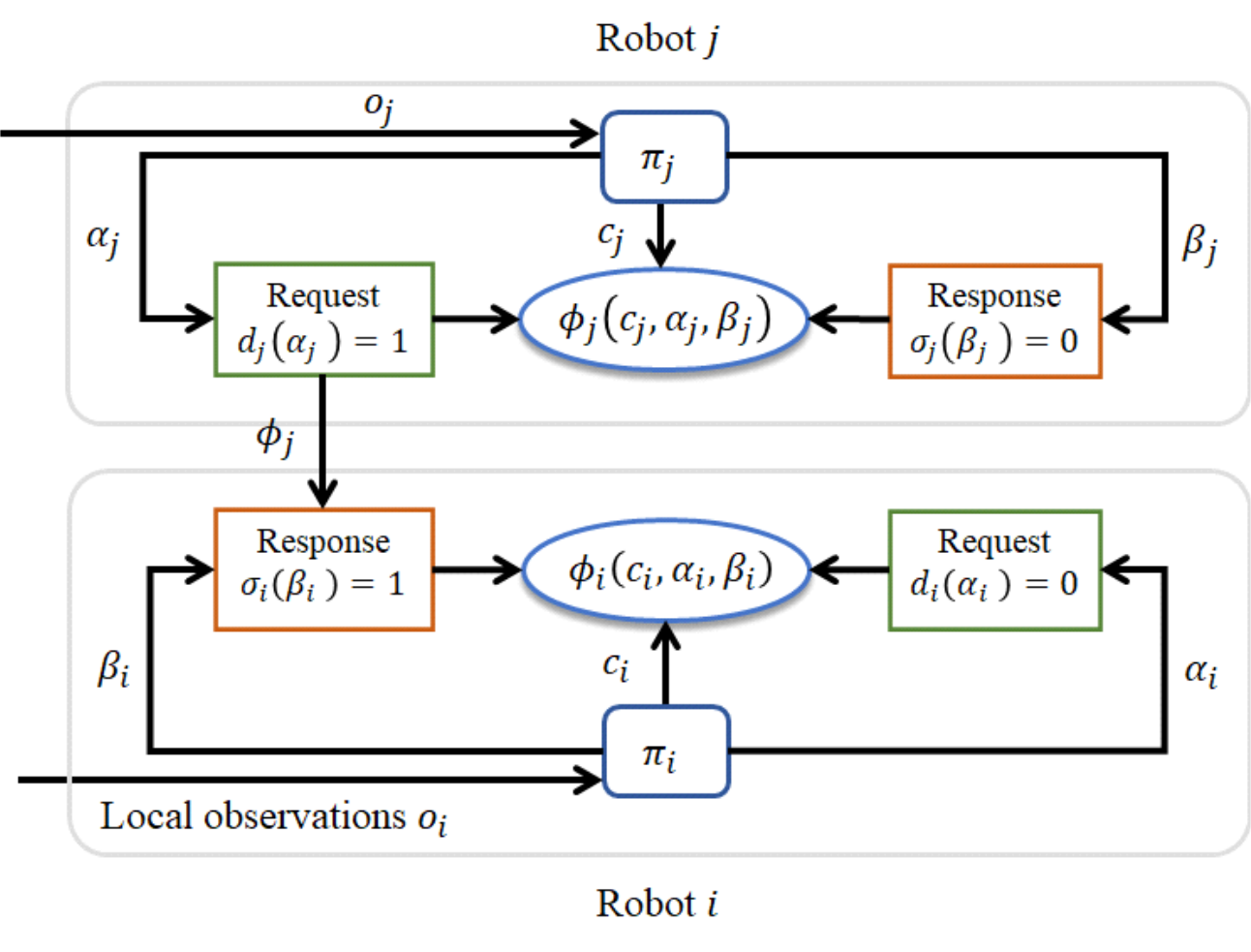}
\vspace{-3mm}
\caption{Example of the communication of the task priority using the proposed distributed policy under local observation. When $d_j(\alpha_j)=1$ and $\sigma_i(\beta_i)=1$, robot $i$ receives $\boldsymbol{\phi}_j$ transmitted by robot $j$.}
\label{communication}
\end{center}
\end{figure}

Using $\alpha_i$ and $\beta_i$ in (\ref{policy}), request signal $d_i$ and response signal $\sigma_i$ are calculated by the event-triggered law (\cite{Baumann2018, Shibata2021}) given by
\begin{eqnarray}
d_i(\alpha_i)&=&
\begin{cases}
1,\ {\rm if}\ \alpha_i>0.5 \ \& \ \|\textit{\textbf{v}}_{l_i^{\ast}}\|_2=0 \\
0,\ {\rm otherwise}
\end{cases}, \label{yousei} \\
\sigma_i(\beta_i)&=&
\begin{cases}
1,\ {\rm if}\ \beta_i>0.5 \ \& \ \|\textit{\textbf{v}}_{l_i^{\ast}}\|_2=0 \\
0,\ {\rm otherwise}
\end{cases}, \label{outou}
\end{eqnarray}
where robot $i$ can transmit and receive the priority when it cannot move the selected object $l_i^{\ast}$.
Fig. \ref{communication} illustrates the communication of the task priority using our distributed policy under local observation.
Using the triggering law in Eqs. (\ref{yousei}) and (\ref{outou}), robot $i$ can receive $\boldsymbol{\phi}_j$ transmitted by robot $j$ and then reach consensus on the high-priority object using (\ref{priority}).

\vspace{-2mm}
\subsection{Object Selection}
\vspace{-2mm}
This subsection introduces the procedure for the selection of an object based on its priority.
Robot $i$ selects the object with the highest priority among $M$ objects using $l_i^{\ast}=\argmax_{l} \phi_i^l$.
Moreover, we set the priority of the object that has reached close to the desired position using $\phi_i^l\leftarrow 0,\ {\rm if}\ \|\textit{\textbf{z}}_l-\textit{\textbf{z}}_l^{\ast}\|_2<\delta$, where $\delta>0$ represents a threshold to determine whether the object reaches the desired position.

\vspace{-2mm}
\subsection{Reward Design}
\vspace{-2mm}
To transport all the objects to the desired positions as quickly as possible, we designed a reward function given by
\vspace{-3mm}
\begin{eqnarray}
r&=&\sum^M_{l=1}P_l+\lambda \sum^M_{l=1}\|\textit{\textbf{v}}_l\|_2, \label{reward1}\\
P_l&=&
\begin{cases}
1,\ {\rm if}\ \|\textit{\textbf{z}}_l-\textit{\textbf{z}}_l^{\ast}\|_2<\delta \\
0,\ {\rm otherwise}
\end{cases},\nonumber
\label{reward}
\end{eqnarray}
where $\lambda$ is a positive constant.
The first term in (\ref{reward}) aims to transport all the objects to the desired positions, while the second term aims to move as many objects as possible.

\vspace{-2mm}
\subsection{Policy Optimization}
\vspace{-2mm}
In this study, we optimized the multi-agent policies using multi-agent deep deterministic policy gradient (MADDPG) (\cite{Lowe2017}), which is one of the deep actor-critic algorithms for multi-agent systems.

\vspace{-1mm}
A common problem of MARL is that the learning becomes unstable because the variance of the policy gradient becomes large when the number of unobservable agents increases. The MADDPG algorithm addressed this problem using a learning framework called "centralized training and decentralized execution." During training, the weight parameters of the critic networks are optimized through the Q-learning algorithm (\cite{Watkins1992}) using the observations and actions of all the agents. Thus, it can reduce the variances of the Q-value functions. In contrast, the weight parameters of the actor networks are optimized through a policy gradient method using its observations and actions. During execution, the actor networks compute actions under local observations. See \cite{Lowe2017} for the details of the policy optimization steps.

\section{SIMULATION}
\vspace{-2mm}
We conducted multi-object transport simulations using multiple robots to confirm the scalability and versatility of the proposed framework for various numbers of robots and objects and various proportions of heavy and light objects.

\vspace{-2mm}
\subsection{Simulation Setup}
\begin{figure}[!tp]
\vspace{-1mm}
\begin{center}
\includegraphics[width=3cm]{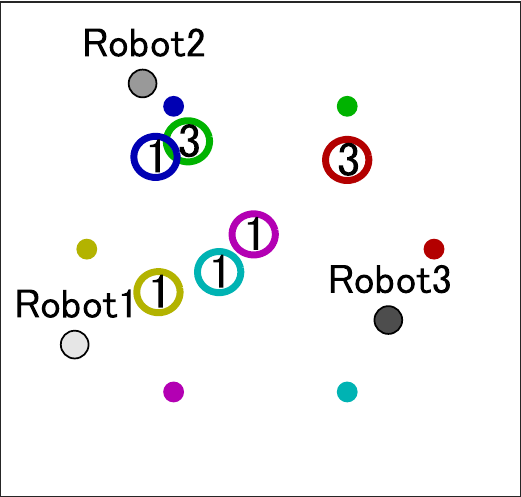}
\vspace{-1mm}
\caption{Simulation scenario. The colored circles, dots, and numbers indicate the objects, their desired positions, and the number of robots required to transport the object, respectively. Robots should transport each object to the desired position with the same color.}
\label{simenv}
\end{center}
\end{figure}

\vspace{-2mm}
We show the simulation scenario in Fig. \ref{simenv}.
We randomly generated the initial positions of the robots and objects in the region $Q:=\{(x,y)\mid2\le x\le 8, 2\le y\le 8\}$.
The desired positions of the objects were evenly arranged on a circumference with a center and radius of $\left[5.0, 5.0\right]^\top$ and 4.0 m, respectively.
We set the load capacity of the robot to 1 kg.

\vspace{-1mm}
During training, we set $K=2$, $N=3$, and $M=6$, while setting the object's mass to 1 or 3 kg with 50 $\%$ probability.
To confirm the scalability of the algorithm, we evaluated $N\in \{3,6\}$ and $M\in \{4,6,8,10\}$.

\vspace{-1mm}
We used the MADDPG code (\cite{maddpg}) and set the simulation parameters as listed in Table 1.

\vspace{-1mm}
We set $k_{\phi}$ in (\ref{priority}) to 0.2 such that the priority changed according to a first-order delay with the time constant of 5 s, which was longer than the selection period.
The threshold $\delta$ was set to 0.05 for the positions of the objects to be controlled within 0.05 m from the desired positions.
We set the weight parameter $\lambda$ in (\ref{reward}) to 3.0$\times 10^2$ such that transporting a different object obtained almost the same reward as locating an object to the desired position.

\vspace{-1mm}
To confirm the effectiveness of our framework, we conducted comparisons through the following methods:
\begin{itemize}
\item \textbf{Nearest}: Each robot selects the nearest object
\item \textbf{One}: Each robot is randomly assigned an object from the $M$ objects
\item \textbf{Local}: Our method without dynamic task priority with global communication by setting $\sigma_i=0$ in (\ref{priority}).
\item \textbf{Nearest-one}: Each robot selects the object closest to its current position. When the robot does not move the object for a specific time $t_s$, the robot picks the same object as the robot, unable to carry the load for the longest time. We set $t_s=1.0$ s for all the robots.
\item \textbf{No-com}: \textbf{Local} method under local observations without the task priorities.
\item \textbf{No-dynamics}: \textbf{No-com} method without the dynamics of the task priority by setting ${\phi}_i^l=c_i^l$ ($l\in \mathcal{N}_i^{\rm Load}$).
\end{itemize}

\begin{table}[!bp]
\small
\caption{Simulation parameters}
\label{sim1_condition}
\vspace{-2mm}
\centering
\renewcommand{\arraystretch}{1.0}
\begin{tabular}{ccc}
\hline
Parameter & Value \\
\hline \hline
Selection period [s] & 1.0 \\
Number of steps per episode & 150 \\
Number of episodes& 2.0e5 \\
Number of hidden layers (critic) & 4 \\
Number of hidden layers (actor) & 4 \\
Number of units per layer & 64 \\
Activation function of hidden layers & ReLU \\
Activation function of output layers (critic) & linear \\
Activation function of output layers (actor) & tanh \\
Discount factor & 0.99 \\
Batch size & 1024 \\
\hline
\end{tabular}
\end{table}

To evaluate our approach quantitatively, we used the following measures:
\begin{itemize}
\item Success rate (SR): The ratio of trials to 100 trials, in which robots can transport all the objects to the desired positions within 10 min. We considered 10 min for method \textbf{One} to achieve a 100 $\%$ success rate for various numbers of robots and objects.
\item Transport time (TT) [$s$]: Average time required to move all the objects to the desired positions within 10 min.
\end{itemize}

\vspace{-2mm}
\subsection{Comparisons of Training Performance}
\vspace{-2mm}
We evaluated the effects of dynamic task priority and communication on the training performance by comparing our framework with methods \textbf{Local}, \textbf{No-com}, and \textbf{No-dynamics}.
For each method, we repeated the training three times.

Fig. \ref{reward} shows the cumulative rewards of the first and second terms in (\ref{reward}), which are denoted as $R_1$ and $R_2$, respectively.
When applying method \textbf{No-dynamics}, we confirmed the occurrences of chattering where the robots travel back and forth between different objects.
As a result, this method made $R_1$ achieve smaller values compared to those in other methods.

Method \textbf{Local} achieves slightly higher $R_1$ and $R_2$ values than method \textbf{No-com}.
Therefore, training the policy with the priority of neighboring robots improves the training performance.
Moreover, the proposed framework achieves higher values than the other methods.
These results indicate that the dynamic task priority with global communication in (\ref{priority}) has a greater impact on the training performance of our framework than the local communication of task priorities.

\vspace{-2mm}
\subsection{Emergence of Cooperative and Independent Actions}
\vspace{-2mm}
We confirmed the emergence of cooperative and independent actions when applying the proposed framework.
We show trajectories and communication occurrences when applying the framework in Fig. \ref{result}.

At the initial stage, robots 1 and 3 transport different objects, while robot 2 cannot move the object, which requires three robots to transport, as shown in Fig. \ref{resulta}.
To prevent this situation, robot 2 transmits its priority to other robots, as shown in Fig. \ref{resultb}.
While robots 1 and 3 receive the priority of robot 2, their priorities gradually approach that of robot 2, as shown in Figs. 7(a) - (c).
Once the priority of the green object is the highest for the three robots, the object is transported to the desired position, as shown in Fig. \ref{resultc}.
Fig. \ref{phi70} shows that the priority of the red object is the highest in the corresponding period for all the robots, which transport the object to the desired position, as shown in Fig. \ref{resultd}.

Finally, two objects remain to be transported by three robots, as shown in Fig. \ref{resulte}.
While three robots transport the light-blue object according to the priority in Fig. \ref{phi90}, the priority of the blue object is the highest for robot 3, as shown in Fig. \ref{phi110}.
Hence, robot 3 moves the blue object, and the two objects can be transported to the desired positions, as shown in Fig. \ref{resultf}.

Overall, our framework can balance cooperative and independent actions by determining the timing of priority communication.

\begin{figure}[!tp]
\centering
\subfigure[First term]{
\includegraphics[width=3.5cm]{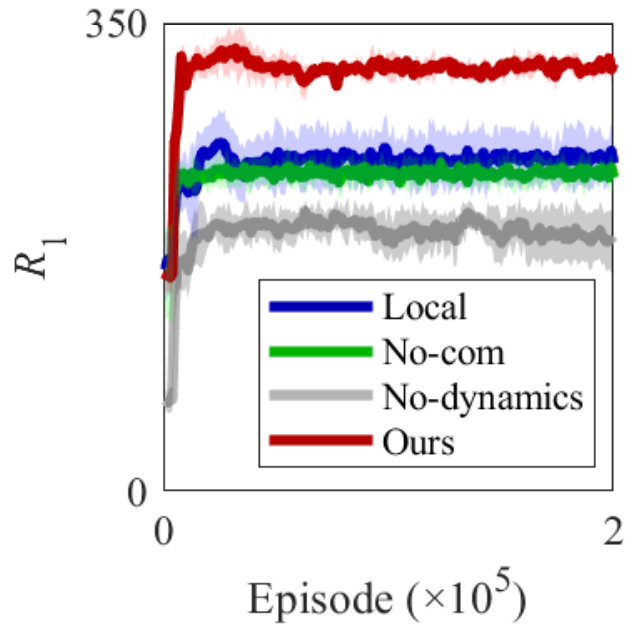}
\label{reward1}}
\subfigure[Second term]{
\includegraphics[width=3.5cm]{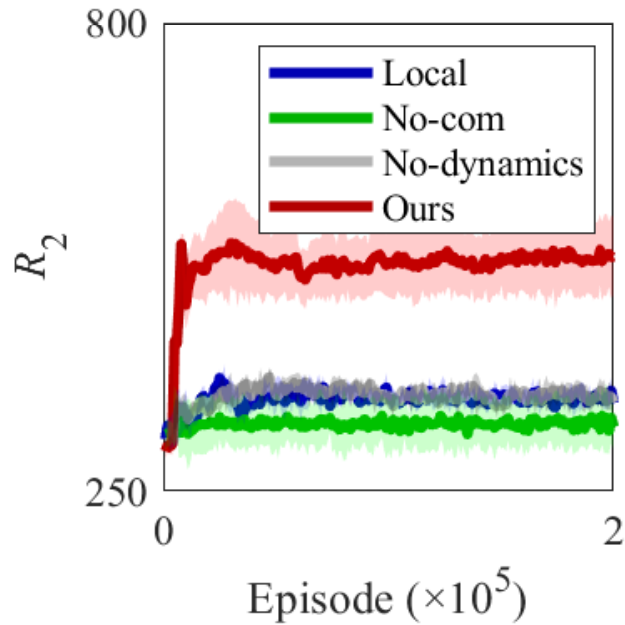}
\label{reward2}}
\vspace{-2mm}
\caption{Cumulative rewards of various evaluated methods.}
\label{reward}
\end{figure}

\begin{figure}[!tp]
\centering
\subfigure[0 - 20 s]{
\includegraphics[width=2cm]{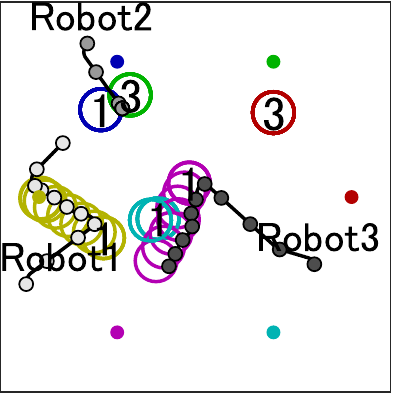}
\label{resulta}}
\hspace{1mm}
\subfigure[20 - 40 s]{
\includegraphics[width=2cm]{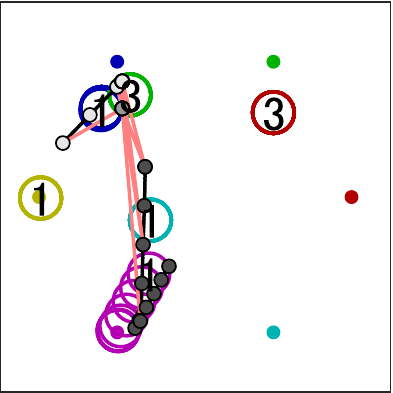}
\label{resultb}}
\hspace{1mm}
\subfigure[40 - 60 s]{
\includegraphics[width=2cm]{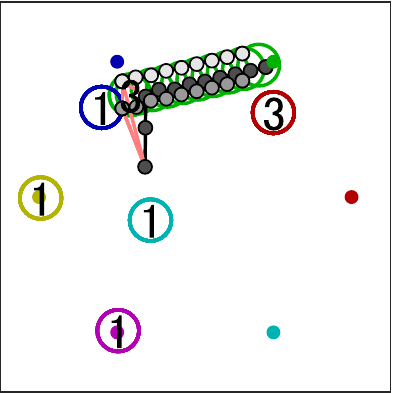}
\label{resultc}}
\hspace{1mm}
\subfigure[60 - 80 s]{
\includegraphics[width=2cm]{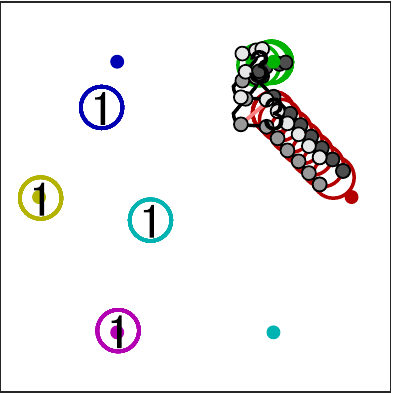}
\label{resultd}}
\hspace{1mm}
\subfigure[80 - 100 s]{
\includegraphics[width=2cm]{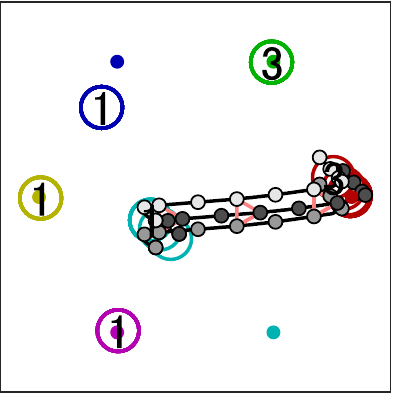}
\label{resulte}}
\hspace{1mm}
\subfigure[100 - 120 s]{
\includegraphics[width=2cm]{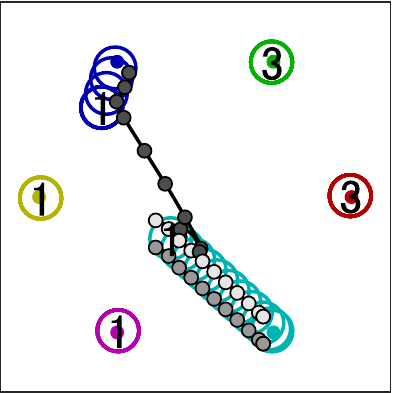}
\label{resultf}}
\vspace{-3mm}
\caption{Trajectories and communication occurrences. We offset the overlapping robot trajectories for clarity. The black and red lines show the trajectories of the robots and occurrences of priority communication.}
\label{result}
\end{figure}

\vspace{-2mm}
\begin{figure}[!tp]
\centering
\subfigure[10 s]{
\includegraphics[width=2.5cm]{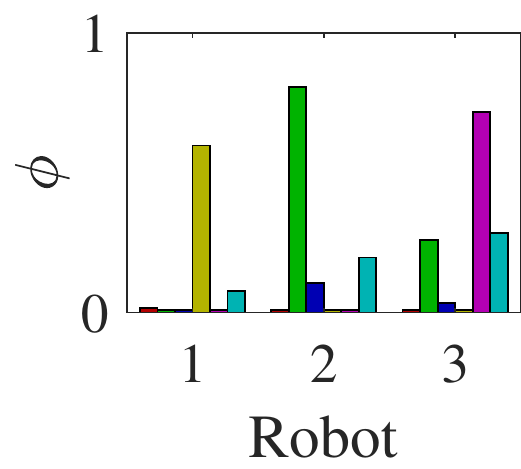}
\label{phi10}}
\subfigure[30 s]{
\includegraphics[width=2.5cm]{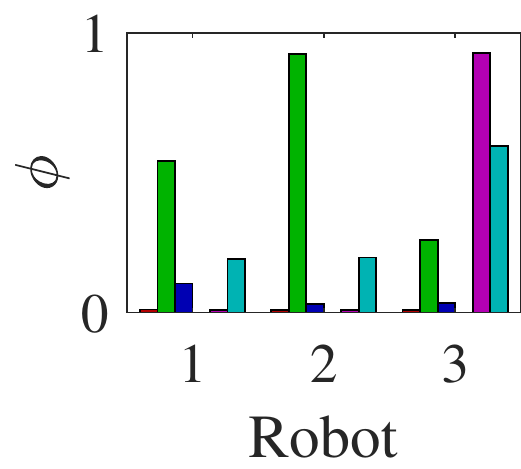}
\label{phi30}}
\subfigure[50 s]{
\includegraphics[width=2.5cm]{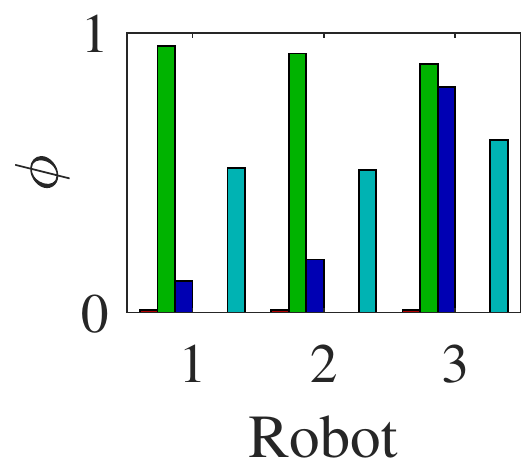}
\label{phi50}}
\subfigure[70 s]{
\includegraphics[width=2.5cm]{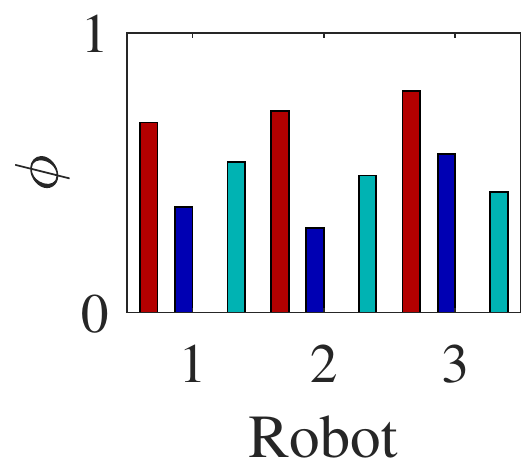}
\label{phi70}}
\subfigure[90 s]{
\includegraphics[width=2.5cm]{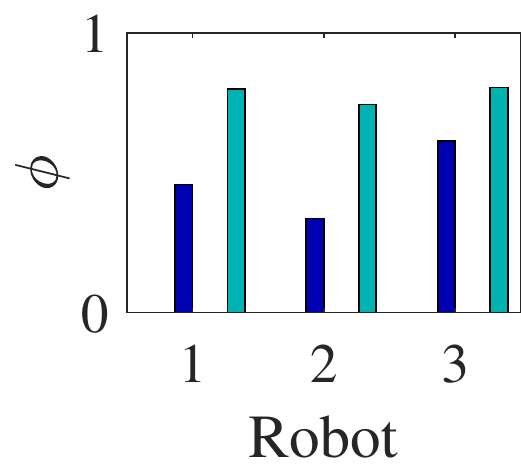}
\label{phi90}}
\subfigure[110 s]{
\includegraphics[width=2.5cm]{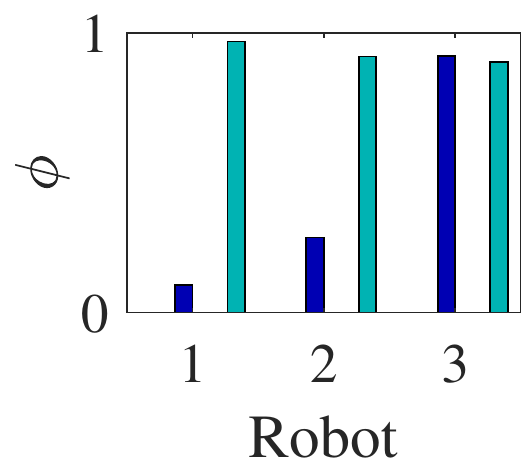}
\label{phi110}}
\vspace{-3mm}
\caption{Task priorities per robot. The colors correspond to those of the objects shown in Fig. \ref{result}.}
\label{phi}
\end{figure}

\vspace{-2mm}
\subsection{Scalability Analysis}
\vspace{-2mm}
We evaluated the success rate and transport time when using our framework and other methods for various numbers of robots and objects.

Table \ref{table:scalability} shows the quantitative results for various numbers of robots and objects when applying each method.
Methods \textbf{Nearest} and \textbf{Local} cannot achieve a 100 $\%$ success rate for various numbers of robots and objects.
In contrast, \textbf{One}, \textbf{Nearest-one}, and \textbf{Ours} can achieve a 100 $\%$ success rate for various numbers of robots and objects.

When applying Method \textbf{One}, the transport time is the longest because all the robots select the same object. To confirm the effectiveness of our method, we evaluated the average time $t_a$ for carrying two or more objects simultaneously when applying each method to $(N, M)=(6, 10)$ for 100 trials. Our method achieves $t_a=6.0\times 10^1$ while Method \textbf{Nearest-one} achieves $t_a=4.9\times 10^1$. The results indicate that our method can promote more independent actions compared to Method \textbf{Nearest-one}.

Overall, compared to other methods, our framework can reduce the transport time while transporting all the objects to the desired positions for various numbers of robots and objects.

\begin{table}[!tp]
\caption{Quantitative results for various numbers of robots and objects}
\vspace{-2mm}
  \label{table:scalability}
  \centering
  \renewcommand{\arraystretch}{1.0}
  \small
  \begin{tabular}{ccccccc}
    \hline
    \multirow{2}{*}{($N,M$)} \hspace{-2mm} & \multirow{2}{*}{Metrics} \hspace{-2mm} & \multirow{2}{*}{Nearest} \hspace{-2mm} & \multirow{2}{*}{One} \hspace{-2mm} & \multirow{2}{*}{Local} \hspace{-2mm} & Nearest \hspace{-2mm} & \multirow{2}{*}{\textbf{Ours}} \hspace{-2mm} \\
    & & & & & -one & \\
    \hline \hline
    \multirow{2}{*}{(3,4)} & SR & 0.59 & \textbf{1.0} & 0.93 & \textbf{1.0} & \textbf{1.0} \\ 
     & TT ($\times 10^2$) & \textbf{1.1} & 1.4 & 1.3 & 1.2 & 1.2 \\ 
    \hline
    \multirow{2}{*}{(3,6)} & SR & 0.34 & \textbf{1.0} & 0.85 & \textbf{1.0} & \textbf{1.0} \\ 
     & TT ($\times 10^2$) & \textbf{1.7} & 2.0 & 2.2 & 1.9 & 1.8 \\ 
    \hline
    \multirow{2}{*}{(3,8)} & SR & 0.17 & \textbf{1.0} & 0.67 & \textbf{1.0} & \textbf{1.0} \\ 
     & TT ($\times 10^2$) & \textbf{2.2} & 2.7 & 2.8 & 2.5 & 2.4 \\ 
    \hline 
    \multirow{2}{*}{(3,10)} & SR & 0.07 & \textbf{1.0} & 0.54 & \textbf{1.0} & \textbf{1.0} \\ 
     & TT ($\times 10^2$) & \textbf{2.5} & 3.4 & 3.5 & 3.1 & 3.0 \\ 
    \hline 
    \multirow{2}{*}{(6,4)} & SR & 0.97 & \textbf{1.0} & \textbf{1.0} & \textbf{1.0} & \textbf{1.0} \\ 
     & TT ($\times 10^2$) & 0.99 & 1.3 & 0.91 & 0.88 & \textbf{0.86} \\ 
    \hline
    \multirow{2}{*}{(6,6)} & SR & 0.88 & \textbf{1.0} & 0.98 & \textbf{1.0} & \textbf{1.0} \\ 
     & TT ($\times 10^2$) & 1.5 & 2.0 & \textbf{1.4} & \textbf{1.4} & \textbf{1.4} \\ 
    \hline
    \multirow{2}{*}{(6,8)} & SR & 0.82 & \textbf{1.0} & 0.88 & \textbf{1.0} & \textbf{1.0} \\ 
     & TT ($\times 10^2$) & 1.9 & 2.7 & 1.9 & 1.9 & \textbf{1.8} \\ 
    \hline 
    \multirow{2}{*}{(6,10)} & SR & 0.74 & \textbf{1.0} & 0.72 & \textbf{1.0} & \textbf{1.0} \\ 
     & TT ($\times 10^2$) & 2.6 & 3.6 & 2.5 & 2.4 & \textbf{2.2} \\ 
    \hline                     
  \end{tabular}
\end{table}

\vspace{-2mm}
\subsection{Versatility Analysis by Varying Proportion of Heavy and Light Objects}
\vspace{-2mm}
Additionally, we verified the versatility of our framework by varying the proportion of heavy and light objects.
We set $N=6$ and $M=10$ while setting the mass of the objects to 1 or 3 kg.
We evaluated each method by generating 3 kg objects with probabilities of 0 $\%$, 25 $\%$, 50 $\%$, 75 $\%$, and 100 $\%$.

\vspace{-1mm}
Table \ref{table:versatility} shows the quantitative results for various proportions of heavy and light objects.
When applying methods \textbf{Nearest} and \textbf{Local}, the success rate becomes lower with the increasing proportion of heavy objects.
In contrast, \textbf{One}, \textbf{Nearest-one}, and \textbf{Ours} can achieve a 100 $\%$ success rate for various proportions of heavy objects.

\vspace{-1mm}
Method \textbf{One} increases the transport time compared with \textbf{Nearest-one} and our methods because all the robots select a common object regardless of its weight.
Moreover, our framework achieves a lower transportation time than method \textbf{Nearest-one} for various proportions of heavy objects because our framework can promote more independent actions than Method \textbf{Nearest-one} as discussed in subsection 5.4.

\vspace{-1mm}
Overall, compared to other methods, our framework can reduce the transport time while transporting all the objects to their desired positions when handling objects with various weights.

\begin{table}[!tp]
\caption{Quantitative results for various proportions of heavy and light objects. $P$ represents the proportion of heavy objects.}
\vspace{-2mm}
  \label{table:versatility}
  \centering
  \renewcommand{\arraystretch}{1.0}
  \small
  \begin{tabular}{ccccccc}
    \hline
    \multirow{2}{*}{$P$} \hspace{-2mm} & \multirow{2}{*}{Metrics} \hspace{-2mm} & \multirow{2}{*}{Nearest} \hspace{-2mm} & \multirow{2}{*}{One} \hspace{-2mm} & \multirow{2}{*}{Local} \hspace{-2mm} & Nearest \hspace{-2mm} & \multirow{2}{*}{\textbf{Ours}} \hspace{-2mm} \\
    & & & & & -one & \\
    \hline \hline
    \multirow{2}{*}{0.0} & SR & \textbf{1.0} & \textbf{1.0} & \textbf{1.0} & \textbf{1.0} & \textbf{1.0} \\ 
     & TT ($\times 10^2$) & 1.2 & 3.3 & 1.1 & 1.2 & \textbf{1.0} \\ 
    \hline
    \multirow{2}{*}{0.25} & SR & 0.94 & \textbf{1.0} & 0.98 & \textbf{1.0} & \textbf{1.0} \\ 
     & TT ($\times 10^2$) & 1.9 & 3.5 & 1.7 & 1.8 & \textbf{1.6} \\ 
    \hline
    \multirow{2}{*}{0.5} & SR & 0.7 & \textbf{1.0} & 0.76 & \textbf{1.0} & \textbf{1.0} \\ 
     & TT ($\times 10^2$) & 2.5 & 3.5 & 2.4 & 2.4 & \textbf{2.2} \\ 
    \hline 
    \multirow{2}{*}{0.75} & SR & 0.6 & \textbf{1.0} & 0.36 & \textbf{1.0} & \textbf{1.0} \\ 
     & TT ($\times 10^2$) & 3.1 & 3.7 & 3.2 & 2.9 & \textbf{2.8} \\ 
    \hline 
    \multirow{2}{*}{1.0} & SR & 0.29 & \textbf{1.0} & 0.14 & \textbf{1.0} & \textbf{1.0} \\ 
     & TT ($\times 10^2$) & 3.5 & 3.8 & 3.7 & 3.4 & \textbf{3.3} \\ 
    \hline                   
  \end{tabular}
\end{table}

\vspace{-2.5mm}

\section{CONCLUSIONS}
\vspace{-2.5mm}
We propose a learning framework that can handle scenarios for various numbers of robots and objects with different and unknown weights. The distributed policy model builds consensus on the high-priority object under local observations, thus balancing the cooperative and independent actions. Therefore, compared to other methods, our framework can reduce the transport time while transporting all the objects to their desired positions for various numbers of robots and objects with different and unknown weights.

\vspace{-1mm}
In the present study, we assume that each robot knows the positions of all the objects. Therefore, we may combine our framework with recurrent MARL models (\cite{REWang2020}) and confirm its effectiveness under partial observations with several unknown object positions. Furthermore, our framework requires global communication between robots. Therefore, we should decentralize the communication structure using techniques such as an attentional communication channel (\cite{Zhai2021}).

\vspace{-1mm}
In future work, we will validate our framework through experiments on real robots. In addition, we intend to apply our framework to allocation problems involving a team of heterogeneous robots.


\bibliography{ifacconf}             
                                                   







\end{document}